\documentclass[twoside,11pt]{article}

\usepackage{jmlr2e}

\usepackage[scaled=0.92]{helvet}

\usepackage{graphicx}
\usepackage{subfigure}
\usepackage{framed}
\usepackage{mdwlist}
\usepackage{bm}

\usepackage{url}
\usepackage{amsmath}
\usepackage{sectsty}
\usepackage{amsfonts}

\usepackage{algorithm}
\usepackage{algorithmic}

\newcommand{\mysec}[1]{Section~\ref{sec:#1}}

\newcommand{\myeq}[1]{Eq.~\ref{eq:#1}}

\newcommand{\myfig}[1]{Figure~\ref{fig:#1}}

\newcommand{\E}[2]{\textrm{E}_{#1}\left[#2\right]}
\newcommand{\df}{{\rm d}}
\newcommand{\g}{\,\vert\,}

\newcommand{\expect}[2]{\mathbb{E}_{#1}\left[ #2 \right]}

\jmlrheading{}{2013}{}{00/00}{00/00}{Chong Wang and David M. Blei}

\ShortHeadings{Variational Inference in Nonconjugate Models}{Wang and Blei}
\firstpageno{1}

\begin{document}

\title{Variational Inference in Nonconjugate Models}

\author{\name Chong Wang \email chongw@cs.cmu.edu \\
       \addr Machine Learning Department\\
       Carnegie Mellon University\\
       Pittsburgh, PA, 15213, USA
       \AND
       \name David M. Blei \email blei@cs.princeton.edu \\
       \addr Department of Computer Science\\
       Princeton University\\
       Princeton, NJ, 08540, USA
}

\editor{}

\maketitle

\begin{abstract}

  Mean-field variational methods are widely used for approximate
  posterior inference in many probabilistic models.  In a typical
  application, mean-field methods approximately compute the posterior
  with a coordinate-ascent optimization algorithm.  When the model is
  conditionally conjugate, the coordinate updates are easily derived
  and in closed form. However, many models of interest---like the
  correlated topic model and Bayesian logistic regression---are
  nonconjuate. In these models, mean-field methods cannot be directly
  applied and practitioners have had to develop variational algorithms
  on a case-by-case basis.  In this paper, we develop two generic
  methods for nonconjugate models, Laplace variational inference and
  delta method variational inference.  Our methods have several
  advantages: they allow for easily derived variational algorithms
  with a wide class of nonconjugate models; they extend and unify some
  of the existing algorithms that have been derived for specific
  models; and they work well on real-world datasets. We studied our
  methods on the correlated topic model, Bayesian logistic regression,
  and hierarchical Bayesian logistic regression.

\end{abstract}


\section{Introduction}

Mean-field variational inference lets us efficiently approximate
posterior distributions in complex probabilistic
models~\citep{Jordan:1999,Wainwright:2008}.  Applications of
variational inference are widespread.  As examples, it has been
applied to Bayesian mixtures~\citep{Attias:2000,Adrian:2001},
factorial models~\citep{Ghahramani:1997}, and probabilistic topic
models~\citep{Blei:2003b}.

The basic idea behind mean-field inference is the following. First
define a family of distributions over the hidden variables where each
variable is assumed independent and governed by its own parameter.
Then fit those parameters so that the resulting distribution is close to
the conditional distribution of the hidden variables given the
observations. Closeness is measured with the Kullback-Leibler divergence.
Inference becomes optimization.

In many settings this approach can be used as a ``black box''
technique.  In particular, this is possible when we can easily compute
the conditional distribution of each hidden variable given all of the
other variables, both hidden and observed. (This class contains the
models mentioned above.) For such models, which are called
\textit{conditionally conjugate} models, it is easy to derive a
coordinate ascent algorithm that optimizes the parameters of the
variational distribution~\citep{Beal:2003,Bishop:2006}.  This is the
principle behind software tools like VIBES~\citep{Bishop:2003} and
Infer.NET~\citep{InferNET10}, which allow practitioners to define
models of their data and immediately approximate the corresponding
posterior with variational inference.


Many models of interest, however, do not enjoy the properties required
to take advantage of this easily derived algorithm.  Such
\textit{nonconjugate} models\footnote{ \citet{Carlin:1991} coined the
  term ``nonconjugate model'' to describe a model that does not enjoy
  full conditional conjugacy.}  include Bayesian logistic
regression~\citep{Jaakkola:1997a}, Bayesian generalized linear
models~\citep{Wells:2001}, discrete choice models~\citep{Braun:2010},
Bayesian item response models~\citep{Clinton:2004,Fox:2010}, and
nonconjugate topic models~\citep{Blei:2006d,Blei:2007}. Using
variational inference in these settings requires algorithms tailored
to the specific model at hand.  Researchers have developed a variety
of strategies for a variety of models, including
approximations~\citep{Braun:2010,Ahmed:2007}, alternative
bounds~\citep{Jaakkola:1997a,Blei:2006d,Blei:2007,Khan:2010}, and
numerical quadrature~\citep{Honkela:2004}.

In this paper we develop two approaches to mean-field variational
inference for a large class of nonconjugate models.  First we develop
\textit{Laplace variational inference}.  This approach embeds Laplace
approximations---an approximation technique for continuous
distributions~\citep{Tierney:1989,MacKay:1992}---within a variational
optimization algorithm.  We then develop \textit{delta method
  variational inference}.  This approach optimizes a Taylor
approximation of the variational objective. The details of the
algorithm depend on how the approximation is formed. Formed one way,
it gives an alternative interpretation of Laplace variational
inference.  Formed another way, it is equivalent to using a
multivariate delta approximation~\citep{Bickel:2007} of the
variational objective.

Our methods are generic. Given a model, they can be derived nearly as
easily as traditional coordinate-ascent inference.  Unlike traditional
inference, however, they place fewer conditions on the model,
conditions that are less restrictive than conditional conjugacy. Our
methods significantly expand the class of models for which mean-field
variational inference can be easily applied.

We studied our algorithms with three nonconjugate models: Bayesian
logistic regression~\citep{Jaakkola:1997a}, hierarchical logistic
regression~\citep{Gelman:2007}, and the correlated topic
model~\citep{Blei:2007}. We found that our methods give better results
than those obtained through special-purpose techniques. Further, we
found that Laplace variational inference usually outperforms delta
method variational inference, both in terms of computation time and
the fidelity of the approximate posterior.

\paragraph{Related work.} We have described the various approaches
that researchers have developed for specific models.  There have been
other efforts to examine generic variational inference in nonconjugate
models. \cite{Paisley:2012} proposed a variational inference approach
using stochastic search for nonconjugate models, approximating the
intractable integrals with Monte Carlo methods.  \cite{Gershman:2012a}
proposed a nonparametric variational inference algorithm, which can be
applied to nonconjugate models. \citet{Knowles:2011} presented a
message passing algorithm for nonconjugate models, which has been
implemented in Infer.NET~\citep{InferNET10}; their technique applies
to a subset of models described in this paper.\footnote{It may be
  generalizable to the full set. However, one must determine how to
  compute the required expectations.}

Laplace approximations have been used in approximate inference in more
complex models, though not in the context of mean-field variational
inference. \cite{Smola:2003} used them to approximate the
difficult-to-compute moments in expectation
propagation~\citep{Minka:2001}. \cite{Rue:2009} used them for
inference in latent Gaussian models.  Here we want to use them for
variational inference, in a method that can be applied to a wider
range of nonconjugate models.

Finally, we note that the delta method was first used in variational
inference by~\cite{Braun:2010} in the context of the discrete choice
model. Our method generalizes their approach.

\paragraph{Organization of this paper.}  In
\mysec{nonconjugate-models} we review mean-field variational inference
and define the class of nonconjugate models to which our algorithms
apply.  In \mysec{inference}, we derive Laplace and delta-method
variational inference and present our full algorithm for nonconjugate
inference.  In \mysec{example}, we show how to use our generic method
on several example models and in \mysec{experiments} we study its
performance on these models.  In \mysec{discussion}, we summarize and
discuss this work.

\section{Variational Inference and a Class of Nonconjugate Models}
\label{sec:nonconjugate-models}

We consider a generic model with observations $x$ and hidden variables
$\theta$ and $z$,
\begin{align}\label{eq:joint}
  p(\theta, z, x) = p(x|z) p(z|\theta) p(\theta).
\end{align}
The distinction between the two hidden variables will be made clear
below.




The inference problem is to compute the posterior, the conditional
distribution of $\theta$ and $z$ given $x$,
\begin{equation*}
  p(\theta, z|x) = \frac{p(\theta, z, x)}{\int p(\theta, z,
    x) dz d\theta}.
\end{equation*}
This is intractable for many models because the denominator is
difficult to compute; we must approximate the distribution.  In
variational inference, we approximate the posterior by positing a
simple family of distributions over the latent variables $q(\theta,
z)$ and then finding the member of that family which minimizes the
Kullback-Leibler (KL) divergence to the true
posterior~\citep{Jordan:1999,Wainwright:2008}.\footnote{In this paper,
  we focus on mean-field variational inference where we minimize the
  KL divergence to the posterior.  We note that there are other kinds
  of variational inference, with more structured variational
  distributions or with alternative objective
  functions~\citep{Wainwright:2008,Barber:2012}. In this paper, we use
  ``variational inference'' to indicate mean-field variational
  inference that minimizes the KL divergence.}

In this section we review variational inference and discuss mean-field
variational inference for the class of conditionally conjugate models.
We then define a wider class of nonconjugate models for which
mean-field variational inference is not as easily applied.  In the
next section, we derive algorithms for performing mean-field
variational inference in this larger class of models.

\subsection{Mean-field Variational Inference}

Mean-field variational inference is simplest and most widely used variational
inference method.  In mean-field variational inference we posit a fully
factorized variational family,
\begin{align}
  \label{eq:mean-field-family}
  q(\theta, z) = q(\theta) q(z).
\end{align}
In this family of distributions the variables are independent and each
is governed by its own distribution.  This family usually does not
contain the posterior, where $\theta$ and $z$ are dependent. However,
it is very flexible---it can capture any set of marginals of the
hidden variables.

Under the standard variational theory, minimizing the KL divergence
between $q(\theta, z)$ and the posterior $p(\theta,z|x)$ is equivalent
to maximizing a lower bound of the log marginal likelihood of the
observed data $x$. We obtain this bound with Jensen's inequality,
\begin{align}
  \log p(x) &= \textstyle \log \int p(\theta, z, x) \df z \df \theta
  \nonumber \\
  &\geq \expect{q}{\log p(\theta, z, x)} - \expect{q}{\log q(\theta,
    z)}
  \nonumber \\
  &\triangleq \mathcal{L}(q), \label{eq:lower-bound}
\end{align}
where $\expect{q} {\cdot}$ is the expectation taken with respect to
$q$ and note the second term is the entropy of $q$.  We call ${\cal
  L}(q)$ the variational objective.

Setting ${\partial \mathcal{L}(q)}/ {\partial q} = 0$ shows that the optimal
solution satisfies the following,
\begin{align}
  q^*(\theta) &\propto \textstyle \exp\left\{\expect{q(z)}{\log
      p(z|\theta)p(\theta)} \right\} \label{eq:q-theta}  \\
  q^*(z)  & \propto \textstyle \exp\left\{\expect{q(\theta)} {\log
      p(x|z)p(z|\theta)} \right\} \label{eq:q-z}.
\end{align}
Here we have combined the optimal conditions from~\cite{Bishop:2006}
with the particular factorization of \myeq{joint}.  Note that the
variational objective usually contains many local optima.

These conditions lead to the traditional coordinate ascent algorithm
for variational inference. It iterates between holding $q(z)$ fixed to
update $q(\theta)$ from \myeq{q-theta} and holding $q(\theta)$ fixed
to update $q(z)$ from \myeq{q-z}.  This converges to a local optimum
of the variational objective~\citep{Bishop:2006}.

When all the nodes in a model are \textit{conditionally conjugate},
the coordinate updates of \myeq{q-theta} and \myeq{q-z} are available
in closed form.  A node is conditionally conjugate when its
conditional distribution given its Markov blanket (i.e., the set of
random variables that it is dependent on in the posterior) is in the
same family as its conditional distribution given its parents (i.e.,
its factor in the joint distribution).  For example, in \myeq{joint}
suppose the factor $p(\theta)$ is a Dirichlet and both factors
$p(z\g\theta)$ and $p(x\g z)$ are multinomials.  This means that the
conditional $p(\theta|z)$ is also a Dirichlet and the conditional $p(z
\g x, \theta)$ is also a multinomial. This model, which is latent
Dirichlet allocation~\citep{Blei:2003b}, is conditionally
conjugate. Many applications of variational inference have been
developed for this type of model~\citep{Bishop:1999, Attias:2000,
  Beal:2003}.


However, if there exists any node in the model that is not
conditionally conjugate then this coordinate ascent algorithm is
not available.  That setting arises in many practical models and
does not permit closed-form updates or easy calculation of the
variational objective.  We will develop generic variational
inference algorithms for a wide class of nonconjugate models.
First, we define that class.

\subsection{A Class of Nonconjugate Models}
\label{sec:condition}

We present a wide class of nonconjugate models, still assuming the
factorization of~\myeq{joint}.
\begin{enumerate}
\item We assume that $\theta$ is real-valued and the distribution
  $p(\theta)$ is twice differentiable with respect to $\theta$. If we
  require $\theta>\theta_0$ ($\theta_0$ is a constant), we may define
  a distribution over $\log(\theta-\theta_0)$.  These assumptions
  cover exponential families, such as the Gaussian, Poisson and
  gamma, as well as more complex distributions, such as a
  student-t.

\item We assume the distribution $p(z\g\theta)$ is in the exponential
  family~\citep{Brown:1986},
  \begin{align}\label{eq:z-dist}
    \textstyle p(z\g\theta) = h(z) \exp\left\{\eta(\theta)^\top t(z) -
      a(\eta(\theta))\right\},
  \end{align}
  where $h(z)$ is a function of $z$; $t(z)$ is the sufficient
  statistic; $\eta(\theta)$ is the natural parameter, which is a
  function of the conditioning variables; and $a(\eta(\theta))$ is the
  log partition function.  We also assume that $\eta(\theta)$ is twice
  differentiable; since $\theta$ is real-valued, this is satisfied in
  most statistical models.  Unlike in conjugate models, these
  assumptions do not restrict $p(\theta)$ and $p(z \g \theta)$ to be a
  conjugate pair; the conditional distribution $p(\theta \g z)$ is not
  necessarily in the same family as the prior $p(\theta)$.

\item The distribution $p(x\g z)$ is in the exponential family,
  \begin{align}\label{eq:x-dist}
    \textstyle p(x\g z) = h(x) \exp\left\{t(z)^\top \langle t(x), 1
      \rangle\right\}.
  \end{align}
  We set up this exponential family so that the natural parameter for
  $x$ is all but the last component of $t(z)$ and the last component
  is the negative log normalizer $-a(\cdot)$.  Thus, the distribution
  of $z$ is conjugate to the conditional distribution of $x$; the
  conditional $p(z \g \theta, x)$ is in the same family as $p(z \g
  \theta)$~\citep{Bernardo:1994}.
\end{enumerate}
Our terminology follows these assumptions: $\theta$ is the
\textit{nonconjugate variable}, $z$ is the \textit{conjugate
  variable}, and $x$ is the \textit{observation}.

This class of models is larger than the class of conditionally
conjugate models.  Our expanded class also includes nonconjugate
models like the correlated topic model~\citep{Blei:2007}, dynamic
topic model~\citep{Blei:2006d}, Bayesian logistic
regression~\citep{Jaakkola:1997a,Gelman:2007}, discrete choice
models~\citep{Braun:2010}, Bayesian ideal point
models~\citep{Clinton:2004}, and many others.  Further, the methods we
develop below are easily adapted more complicated graphical models,
those that contain conjugate and nonconjugate variables whose
dependencies are encoded in a directed acyclic graph.  Appendix A
outlines how to adapt our algorithms to this more general case.

\paragraph{Example: Hierarchical language modeling.}  We introduce the
hierarchical language model, a simple example of a nonconjugate model
to help ground our derivation of the general algorithms.  Consider the
problem of unigram language modeling.  We are given a collection of
documents ${\cal D} = x_{1:D}$ where each document $x_d$ is a vector
of word counts, observations from a discrete vocabulary of length
$V$. We model each document with its own distribution over words and
place a Dirichlet prior on that distribution.  This model is used, for
example, in the language modeling approach to information
retrieval~\citep{Croft:2003}.

We want to place a prior on the Dirichlet parameters, a positive
$V$-vector, that govern each document's distribution over terms.
In theory, every exponential family distribution has a conjugate
prior~\citep{Bernardo:1994} and the prior to the Dirichlet is the
multi-gamma distribution~\citep{Kotz:2000}.  However, the
multi-gamma is difficult to work with because its log normalizer
is not easy to compute. As an alternative, we place a log normal
distribution on the Dirichlet parameters.  This is not the
conjugate prior.


The full generative process is as follows:
\begin{enumerate*}
\item Draw log Dirichlet parameters $\theta \sim \mathcal{N}(0, I)$.
\item For each document $d$, $1\leq d \leq D$:
  \begin{enumerate*}
  \item Draw multinomial parameter $z_d \g \theta \sim {\rm
      Dirichlet}(\exp\{\theta\})$.
  \item Draw word counts $x_{d} \sim {\rm Multinomial}(N, z_d)$.
  \end{enumerate*}
\end{enumerate*}
Given a collection of documents, our goal is to compute the
posterior distribution $p(\theta, z_{1:D} \g x_{1:D})$.
Traditional variational or Gibbs sampling methods cannot be
easily used because the normal prior on the parameters $\theta$
is not conjugate to the $\textrm{Dirichlet}(\exp\{\theta\})$
likelihood.

This language model fits into our model class.  In the notation of the
joint distribution of \myeq{joint}, $\theta = \theta$, $z = z_{1:D}$,
and $x = x_{1:D}$.  The per-document multinomial parameters $z$ and
word counts $x$ are conditionally independent given the Dirichlet
parameters $\theta$,
\begin{align*}
  p(z \g \theta) &= \textstyle \prod_d p(z_d \g \theta) \\
       p(x \g z) &= \textstyle \prod_d \prod_n p(x_{dn} \g z_d).
\end{align*}
In this case, the natural parameter $\eta(\theta) =
\exp\{\theta\}$. This model satisfies the assumptions: the log normal
$p(\exp\{\theta\})$ is not conjugate to the Dirichlet $p(z_d \g
\exp\{\theta\})$ but is twice differentiable; the Dirichlet is
conjugate to the multinomial $p(x_d \g z_d)$ and the multinomial is in
the exponential family.

Below we will use various components of the exponential family form of
the Dirichlet:
\begin{align}
  h(z_d) =\textstyle \prod_i z_{di}^{-1}; \ \ t(z_d) = \log z_d; \ \
  a_d(\eta(\theta)) = \sum_i \log\Gamma (\exp\{\theta_i\} -
  \log\Gamma \left(\sum_i\exp\{\theta_i\}\right).\label{eq:example-z}
\end{align}
We will return to this model as a simple running example.

\section{Laplace and Delta Method Variational Inference}
\label{sec:inference}

We have defined a class of nonconjugate models.  Variational inference
is difficult to derive for these models because $p(\theta)$ is not
conjugate to $p(z\g\theta)$.  Specifically, the update in
\myeq{q-theta} does not necessarily have the form of an exponential
family we can work with and it is difficult to use
$\expect{q(\theta)}{\log p(z\g \theta)}$ in the update of \myeq{q-z}.

We will develop two variational inference algorithms for this class:
\textit{Laplace variational inference} and \textit{delta method
  variational inference}.  Both use coordinate ascent to optimize the
variational parameters, iterating between updating $q(\theta)$ and
$q(z)$.  They differ in how they update the variational distribution
of the nonconjugate variable $q(\theta)$.  In Laplace variational
inference, we use Laplace
approximations~\citep{MacKay:1992,Tierney:1989} within the coordinate
ascent updates of \myeq{q-theta} and \myeq{q-z}. In delta method
variational inference, we apply Taylor approximations to approximate
the variational objective in \myeq{lower-bound} and then derive the
corresponding updates.  Different ways of taking the Taylor
approximation lead to different algorithms.  Formed one way, this
recovers the Laplace approximation.  Formed another way, it is
equivalent to using a multivariate delta
approximation~\citep{Bickel:2007} of the variational objective
function.

In both variants, the variational distribution is the mean-field
family in \myeq{mean-field-family}.  The variational distribution of
the nonconjugate variable $q(\theta)$ is a Gaussian; the variational
distribution of the conjugate variable $q(z)$ is in the same family as
$p(z\g\eta(\theta))$.  In Laplace inference, these forms emerge from
the derivation.  In delta method inference, they are assumed.  The
complete variational family is,
\begin{align*}
  q(\theta, z) = q(\theta \g \mu, \Sigma) q(z \g \phi).
\end{align*}
where $(\mu, \Sigma)$ are the parameters for a Gaussian distribution
and $\phi$ is a natural parameter for $z$.  For example, in the
hierarchical language model of \mysec{condition}, $\phi$ is a
collection of $D$ Dirichlet parameters.  We will sometimes supress the
parameters, writing $q(\theta)$ for $q(\theta \g \mu, \Sigma)$.

Our algorithms are coordinate ascent algorithms, where we iterate
between updating the nonconjugate variational distribution $q(\theta)$
and updating the conjugate variational distribution $q(z)$.  In the
subsections below, we derive the update for $q(\theta)$ in each
algorithm.  Then, for both algorithms, we derive the update for
$q(z)$.  The full procedure is described in \mysec{alg} and
\myfig{lp-algorithm}.

\subsection{Laplace Variational Inference}\label{sec:embed-laplace}

We first review the Laplace approximation.  Then we show how to use it
in variational inference.

\paragraph{The Laplace Approximation.} Laplace approximations use a
Gaussian to approximate an intractable density.  Consider
approximating an intractable posterior $p(\theta\g x)$. (There is no
hidden variable $z$ in this set up.) Assume the joint distribution
$p(x,\theta)=p(x\g\theta)p(\theta)$ is easy to compute.  Laplace
approximations use a Taylor approximation around the maximum a
posterior (MAP) point to construct a Gaussian proxy for the posterior.
They are used for continuous distributions.

First, notice the posterior is proportional to the exponentiated log
joint
\[
p(\theta\g x) = \exp \{ \log p(\theta\g x) \} \propto \exp \{\log p(\theta,x) \}.
\]
Let $\hat{\theta}$ be the MAP of $p(\theta\g x)$, found by maximizing $\log
p(\theta,x)$. A Taylor expansion around $\hat{\theta}$ gives
\begin{align}
  \log p(\theta\g x) \approx \log p(\hat{\theta}\g x) +\textstyle \frac{1}{2}
  (\theta-\hat{\theta})^\top
  H(\hat{\theta}) (\theta-\hat{\theta}). \label{eq:laplace}
\end{align}
The term $H(\hat{\theta})$ is the Hessian of $\log p(\theta\g x)$ evaluated at
$\hat{\theta}$, $H(\hat{\theta}) \triangleq  \triangledown^2 \log
p(\theta\g x)
  |_{\theta=\hat{\theta}}$.

In the Taylor expansion of \myeq{laplace}, the first-order term
$(\theta-\hat{\theta})^\top \triangledown \log
p(\theta\g x)|_{\theta=\hat{\theta}} $ equals zero. The reason is that
$\hat{\theta}$ is the maximum of $\log p(\theta\g x)$ and so its
gradient $\triangledown \log p(\theta\g x)|_{\theta=\hat{\theta}}$ is
zero.  Exponentiating \myeq{laplace} gives the approximate Gaussian posterior
\begin{align*}
\textstyle  p(\theta\g x) \approx \frac{1}{C}\exp\left \{-\frac{1}{2}
(\theta-\hat{\theta})^\top \left(-H(\hat{\theta})\right)
(\theta-\hat{\theta})\right\},
\end{align*}
where $C$ is a normalizing constant.  In other words, $p(\theta\g x)$ can be
approximated by
\begin{align} \label{eq:laplace-approximation}
  p(\theta\g x) \approx \mathcal{N} (\hat{\theta}, -H(\hat{\theta})^{-1}).
\end{align}
This is the Laplace approximation.  While powerful, it is difficult to
use in multivariate settings, for example, when there are discrete
hidden variables. Now we describe how we use Laplace approximations as
part of a variational inference algorithm for more complex models.

\paragraph{Laplace updates in variational inference.}  We adapt the
idea behind Laplace approximations to update the variational
distribution $q(\theta)$.  First, we combine the coordinate update
in~\myeq{q-theta} with the exponential family assumption
in~\myeq{z-dist},
\begin{align}\label{eq:q-theta-2}
  \textstyle q(\theta) \propto \exp\left\{ \eta(\theta)^\top
    \expect{q(z)}{t(z)} - a(\eta(\theta)) + \log p(\theta) \right \}.
\end{align}
Define the function $f(\theta)$ to contain the terms inside the
exponent of the update,
\begin{align}
  f(\theta) \triangleq \eta(\theta)^\top \expect{q(z)}{t(z)} -
  a(\eta(\theta)) + \log p(\theta).\label{eq:f-laplace}
\end{align}
The terms of $f(\theta)$ come from the model and involve $q(z)$ or
$\theta$.  Recall that $q(z)$ is in the same exponential family as
$p(z \g \theta)$, $t(z)$ are its sufficient statistics, and $\phi$ is
the variational parameter.  We can compute $\expect{q(z)}{t(z)}$ from
a basic property of the exponential family~\citep{Brown:1986},
\begin{equation*}
  \expect{q(z)}{t(z)} = \nabla a(\phi).
\end{equation*}
Seen another way, $f(\theta) = \expect{q(z)}{\log p(\theta, z)}$.
This function will be important in both Laplace and delta method
inference.

The problem with nonconjugate models is that we cannot update
$q(\theta)$ exactly using \myeq{q-theta-2} because $q(\theta) \propto
\exp\{f(\theta)\}$ cannot be normalized in closed form.  We
approximate the update by taking a second-order Taylor approximation
of $f(\theta)$ around its maximum, following the same logic as from
the original Laplace approximation in \myeq{laplace}.  The Taylor
approximation for $f(\theta)$ around $\hat{\theta}$ is
\begin{align}\label{eq:taylor-f-theta}
  f(\theta) \approx \textstyle f(\hat{\theta}) + \triangledown
  f(\hat{\theta})(\theta-\hat{\theta}) + \frac{1}{2}(\theta-\hat{\theta})^\top
  \triangledown^2 f(\hat{\theta})(\theta-\hat{\theta}),
\end{align}
where $\triangledown^2 f(\hat{\theta})$ is the Hessian matrix
evaluated at $\hat{\theta}$. Now let $\hat{\theta}$ be the value that
maximizes $f(\theta)$.  This implies that $\triangledown
f(\hat{\theta}) =0$ and \myeq{taylor-f-theta} simplifies to
\begin{equation*}
\textstyle  q(\theta) \propto \exp\{f(\theta)\}
  \approx \exp\left\{ f(\hat{\theta}) + \frac{1}{2}(\theta -
  \hat{\theta})^\top \triangledown^2 f(\hat{\theta})
(\theta-\hat{\theta}) \right\}.
\end{equation*}
Thus the approximate update for $q(\theta)$ is to set it to
\begin{align}\label{eq:lap-var}
  q(\theta) \approx \mathcal{N}\left(\hat{\theta}, -\triangledown^2
    f(\hat{\theta})^{-1} \right).
\end{align}
Note we did not assume $q(\theta)$ is Gaussian. Its Gaussian form
stems from the Taylor approximation.

The update in \myeq{lap-var} can be used in a coordinate ascent
algorithm for a nonconjugate model.  We iterate between holding $q(z)$
fixed while updating $q(\theta)$ from \myeq{lap-var}, and holding
$q(\theta)$ fixed while updating $q(z)$.  (We derive the second update
in \mysec{q-z}.)  Each time we update $q(\theta)$ we must use numerical
optimization to obtain $\hat{\theta}$, the optimal value of
$f(\theta)$.

We return to the hierarchical language model of \mysec{condition},
where $\theta$ are the log of the parameters to the Dirichlet
distribution.  Implementing the algorithm to update $q(\theta)$
involves forming $f(\theta)$ for the model at hand and deriving an
algorithm to optimize it.

With the model equations in~\myeq{example-z}, we have
\begin{align}
  f(\theta) = \textstyle \exp(\theta)^\top \expect{q(z)}{t(z)} -
  D\left(\sum_i \log\Gamma (\exp(\theta_i) - \log\Gamma
    \left(\sum_i\exp(\theta_i)\right) \right) -(1/2)\theta^\top \theta.
  \label{eq:unigram-f}
\end{align}
The expected sufficient statistics of the conjugate variable are
\begin{eqnarray*}
  \expect{q(z)}{t(z)} &=& \textstyle \sum_d \expect{q(z_d)}{t(z_d)} \\
  &=& \sum_d\Psi\!\left(\phi_d\right) - \Psi\!\left(\textstyle
    \sum_i \phi_{di}\right),
\end{eqnarray*}
where $\Psi(\cdot)$ is the digamma function, the first derivative of
$\log \Gamma(\cdot)$.  (This function will also arise in the
gradient.)  It is straightforward to use numerical methods, such as
conjugate gradient~\citep{Bertsekas:1999}, to optimize
\myeq{unigram-f}.  We can then use \myeq{lap-var} to update the
nonconjugate variable.

\subsection{Delta Method Variational Inference}
\label{sec:lower-bound}

In Laplace variational inference, the variational distribution
$q(\theta)$ \myeq{lap-var} is solely a function of $\hat{\theta}$, the
maximum of $f(\theta)$ in \myeq{f-laplace}. A natural question is,
would other values of $\theta$ be suitable as well?  To consider such
alternatives, we describe a different technique for variational
inference. We approximate the variational objective $\mathcal{L}$ in
\myeq{lower-bound} and then optimize that approximation.

Again we focus on updating $q(\theta)$ in a coordinate ascent
algorithm and postpone the discussion of updating $q(z)$. We set the
variational distribution $q(\theta)$ to be a Gaussian
$\mathcal{N}(\mu, \Sigma)$, where the parameters are free
variational parameters fit to optimize the variational objective.
(Note that in Laplace inference, this Gaussian family came out of the
derivation.)  We isolate the terms of the objective in
\myeq{lower-bound} related to $q(\theta)$, and we substitute the
exponential family form of $p(z\g \theta)$ in \myeq{z-dist},
\begin{align*}
  \textstyle \mathcal{L}(q(\theta)) =
  \expect{q(\theta)}{\eta(\theta)^\top \expect{q(z)}{t(z)} -
    a(\eta(\theta)) + \log p(\theta)} + \frac{1}{2} \log |\Sigma|.
\end{align*}
The second term comes from the entropy of the Gaussian,
\begin{equation*}
  - \expect{q(\theta)}{\log q(\theta)} = \frac{1}{2} \log |\Sigma|+C,
\end{equation*}
where $C$ is a constant and is excluded from the objective.  The first
term is $\expect{q(\theta)}{f(\theta)}$, where $f(\cdot)$ is the same
as defined for Laplace inference in \myeq{f-laplace}. Thus,
\begin{align*}
  \mathcal{L}(q(\theta)) = \textstyle \expect{q(\theta)}{f(\theta)} +
  \frac{1}{2} \log |\Sigma|.
\end{align*}

We cannot easily compute the expectation in the first term.  So we use
a Taylor approximation of $f(\theta)$ around a chosen value
$\hat{\theta}$ (\myeq{taylor-f-theta}) and then take the expectation,
\begin{align}
\textstyle \mathcal{L}(q(\theta)) \approx f(\hat{\theta}) +  \triangledown
  f(\hat{\theta})^\top(\mu-\hat{\theta}) + & \textstyle
  \frac{1}{2}(\mu-\hat{\theta})^\top
 \triangledown^2 f(\hat{\theta})(\mu-\hat{\theta}) ]] \nonumber \\
 & + \textstyle \frac{1}{2}\left( {\rm Tr}\left\{\triangledown^2 f(\hat{\theta})
  \Sigma\right\} + \log |\Sigma|\right), \label{eq:l-approx}
\end{align}
where ${\rm Tr}(\cdot)$ is the Trace operator.  In the coordinate
update of $q(\theta)$, this is the function we optimize with respect
to its variational parameters $\{\mu, \Sigma\}$.

To fully specify the algorithm we must choose $\hat{\theta}$, the
point around which to approximate $f(\theta)$. We will discuss three
choices.  The first is to set $\hat{\theta}$ to be the maximum of
$f(\theta)$. With this choice, maximizing the approximation in
\myeq{l-approx} gives $\mu=\hat{\theta}$ and $\Sigma = -
\triangledown^2 f(\hat{\theta})^{-1}$.  Notice this is the update
derived in \mysec{embed-laplace}.  We have given a different
derivation of Laplace variational inference.

The second choice is to set $\hat{\theta}$ as the mean of the
variational distribution from the previous iteration of coordinate
ascent. If the prior $p(\theta)$ is Gaussian, this recovers the
updates derived in~\cite{Ahmed:2007} for the correlated topic
model.\footnote{This is an alternative derivation of their algorithm.
  They derived these updates from the perspective of generalized
  mean-field theory~\citep{Xing:2003}.}  In our study, we found this
algorithm did not work well. It did not always converge, possibly due
to the difficulty of choosing an appropriate initial $\hat{\theta}$.

The third choice is to set $\hat{\theta} = \mu$, i.e., the mean of the
variational distribution $q(\theta)$.  With this choice, the variable
around which we center the Taylor approximation becomes part of the
optimization problem.  The objective is
\begin{align}\label{eq:lb-II}
  \mathcal{L}(q(\theta)) &\textstyle \approx f(\mu) + \frac{1}{2} {\rm
    Tr}\left\{\triangledown^2 f(\mu) \Sigma\right\} + \frac{1}{2} \log
  |\Sigma|.
\end{align}
This is the multivariate delta method for evaluating
$\expect{q(\theta)}{f(\theta)}$~\citep{Bickel:2007}.
\textit{Delta method variational inference} optimizes this
objective in the coordinate update of $q(\theta)$ .

In more detail, we first optimize $\mu$ with gradient methods and then
optimize $\Sigma$ in closed form $ \Sigma = - \triangledown^2
f(\mu)^{-1}.$ Note this is more expensive than Laplace variational
inference because optimizing \myeq{lb-II} requires the third
derivative $\triangledown^3 f(\theta)$. \citet{Braun:2010} were the
first to use the delta method in a variational inference algorithm,
developing this technique for the discrete choice model.  If we assume
the prior $p(\theta)$ is Gaussian then we recover their algorithm.
With the ideas presented here, we can now use this strategy in many
models.

We return briefly to the unigram language model.  The delta method
update for $q(\theta)$ optimizes \myeq{lb-II}, using the specific
$f(\cdot)$ found in \myeq{unigram-f}.  While Laplace inference
required the digamma function and $\log \Gamma$ function, delta
method inference will further require the trigamma function.

\subsection{Updating the Conjugate Variable}
\label{sec:q-z}

We derived variational updates for $q(\theta)$ using two methods.  We
now turn to the update for the variational distribution of the
conjugate variable $q(z)$.  We show that both Laplace inference
(\mysec{embed-laplace}) and delta method inference
(\mysec{lower-bound}) lead to the same update. Further, we have
implicitly assumed that $\expect{q(z)}{t(z)}$ in \myeq{f-laplace} is
easy to compute.  We will confirm this as well.

We first derive the update for $q(z)$ when using Laplace inference.
We apply the exponential family form in \myeq{z-dist} to the exact
update of~\myeq{q-z},
\begin{align*}
  \log q(z) = \log p(x\g z) + \log h(z) +
  \expect{q(\theta)}{\eta(\theta)}^\top t(z) + C,
\end{align*}
where $C$ is a constant not depending on $z$. Now we use $p(x\g z)$ from
\myeq{x-dist} to obtain
\begin{align}\label{eq:q-z-exact2}
  \textstyle q(z) \propto h(z) \exp\left
    \{\left(\expect{q(\theta)}{\eta(\theta)} + t(x)\right)^\top
    t(z) \right \},
\end{align}
which is in the same family as $p(z\g\theta)$ in \myeq{z-dist}.  This
is the update for $q(z)$.

Recall that $\eta(\theta)$ maps the nonconjugate variable $\theta$ to
the natural parameter of the conjugate variable $z$.  The update for
$q(z)$ requires computing $\expect{q(\theta)}{\eta(\theta)}$.  For
some models, this expectation is computable.  If not, we can take a
Taylor approximation of $\eta(\theta)$ around the variational
parameter $\mu$,
\begin{align*}
  \eta_i(\theta) \approx \textstyle \eta_i(\mu) + \triangledown
  \eta(\mu)_i^\top(\theta-\mu) + \frac{1}{2}(\theta-\mu)^\top
  \triangledown^2
  \eta_i(\mu)(\theta-\mu),
\end{align*}
where $\eta(\theta)$ is a vector and $i$ indexes the $i$th component.
This requires $\eta(\theta)$ is twice differentiable, which is
satisfied in most models.  Since $q(\theta) = \mathcal{N}\left(\mu,
  \Sigma \right)$, this means that
\begin{align}
  \label{eq:eta-theta-approx}
  \expect{q(\theta)}{\eta_i(\theta)} \approx \textstyle \eta_i(\mu) +
  \frac{1}{2}
  {\rm Tr}\left\{\triangledown^2 \eta_i(\mu)\Sigma\right\}.
\end{align}
(Note that the linear term $\expect{q(\theta)}{\triangledown
  \eta_i(\mu)^T(\theta - \mu)} = 0$.)

Using delta method variational inference to update $q(\theta)$, the
update for $q(z)$ is identical to that in Laplace variational
inference.  We isolate the relevant terms in \myeq{lower-bound},
\begin{align}
  \textstyle \mathcal{L}(q(z)) = &\textstyle \expect{q(z)}{ \log
    p(x\g z) + \log h(z) + \expect{q(\theta)}{\eta(\theta)}^\top t(z)}
  - \expect{q(z)}{\log q(z)}.
  \label{eq:delta-q-z}
\end{align}
Setting the partial gradient $\partial\mathcal{L}(q(z))/\partial q(z)
= 0$ gives the same optimal $q(z)$ of \myeq{q-z}. Computing this
update reduces to the approach for Laplace variational inference in
\myeq{q-z-exact2}.

We return again to the unigram language model with log normal priors
on the Dirichlet parameters.  In this model, we can compute
$\expect{q(\theta)}{\eta(\theta)}$ exactly by using properties of the
log normal,
\begin{align*}
  \expect{q(\theta)}{\eta(\theta)} = \expect{q(\theta)}{\exp\{\theta\}} =
  \exp\{\mu + {\rm diag}(\Sigma)/2\}.
\end{align*}
Recall that $x_d$ are the word counts for document $d$ and note that
it is its own sufficient statistic in a multinomial count model.
Given the calculation of $\expect{q(\theta)}{\eta(\theta)}$ and the
model-specific calculations in ~\myeq{example-z}, the update for
$q(z_d)$ is
\begin{align*}
  q(z_d) = \textstyle {\rm Dirichlet}\left(\exp(\mu + {\rm diag}(\Sigma)/2) +
  x_{d}\right).
\end{align*}
This completes our derivation in the example model.  To implement
nonconjugate inference we need this update for $q(z)$ and the
definition of $f(\cdot)$ in \myeq{unigram-f}.

\subsection{Nonconjugate Variational Inference}
\label{sec:alg}

\begin{figure}[t!]
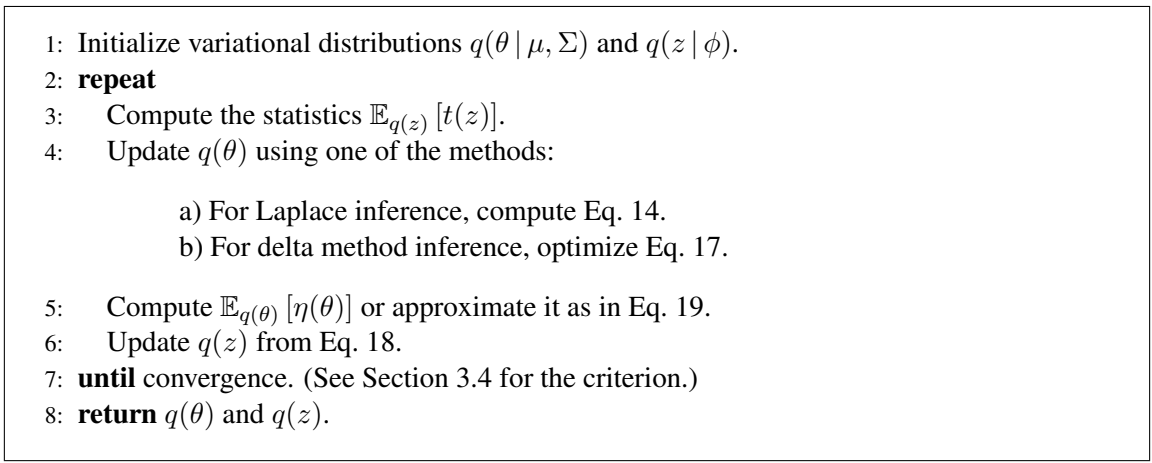

\begin{framed}
\begin{algorithmic}[1]
  \STATE {Initialize variational distributions $q(\theta \g \mu, \Sigma)$ and
    $q(z \g \phi)$.}

\REPEAT \STATE{Compute the statistics
    $\expect{q(z)}{t(z)}$.}

  \STATE { Update $q(\theta)$ using one of the methods:
    \begin{itemize*}
    \item [] a) For Laplace inference, compute \myeq{lap-var}.
    \item [] b) For delta method inference, optimize \myeq{lb-II}.
    \end{itemize*}
}
\STATE{Compute $\expect{q(\theta)}{\eta(\theta)}$ or approximate
it as in \myeq{eta-theta-approx}.}
\STATE {Update $q(z)$ from \myeq{q-z-exact2}.}
\UNTIL{convergence. (See \mysec{alg} for the criterion.)}
\STATE { {\bf return }$q(\theta)$ and $q(z)$.}
\end{algorithmic}
\end{framed}
\caption{Nonconjugate variational inference \label{fig:lp-algorithm}}
\end{figure}

We now present the full algorithm for nonconjugate variational
inference.  In this section, we will be explicit about the
variational parameters.  Recall that the variational distribution of
the nonconjugate variable is a Gaussian $q(\theta \g \mu, \Sigma)$;
the variational distribution of the conjugate variable is $q(z \g
\phi)$, where $\phi$ is a natural parameter in the same family as $p(z
\g \eta(\theta))$.

The algorithm is as follows.  Begin by initializing the variational
parameters.  Iterate between updating $q(\theta)$ and updating $q(z)$
until convergence.  Update $q(\theta)$ by either \myeq{lap-var}
(Laplace inference) or optimizing \myeq{lb-II} (Delta method
inference).  Update $q(z)$ from \myeq{q-z-exact2}.  Assess convergence
by measuring the $L_2$ norm of the mean of the nonconjugate variable,
$\expect{q}{\theta}$.

This algorithm is summarized in \myfig{lp-algorithm}.  In either
Laplace or delta method inference, we have reduced deriving
variational updates for complicated nonconjugate models to mechanical
work---calculating derivatives and calling a numerical optimization
library.  We note that Laplace inference is simpler to derive because
it only requires second derivatives of the function
in~\myeq{f-laplace}; delta method inference requires third
derivatives.  We study the empirical difference between these methods
in \mysec{experiments}.

Our algorithm (in either setting) is based on approximately optimal
coordinate updates for the variational objective, but we cannot
compute that objective.  However, we can compute an approximate
objective at each iteration with the same Taylor approximation used in
the coordinate steps, and this can be monitored as a proxy.  The
approximate objective is
\begin{align}
  \mathcal{L} \approx f(\hat{\theta}) + \triangledown
  f(\hat{\theta})^\top(\mu-\hat{\theta}) +\textstyle & \textstyle
  \frac{1}{2}(\mu-\hat{\theta})^\top
  \triangledown^2 f(\hat{\theta})(\mu-\hat{\theta}) \nonumber \\
  &+ \textstyle \frac{1}{2}\left( {\rm Tr}\left\{\triangledown^2
      f(\hat{\theta}) \Sigma\right\} + \log |\Sigma|\right) -
  \expect{q(z)}{\log q(z)}
  \label{eq:monitor}
\end{align}
where $f(\theta)$ is defined in~\myeq{f-laplace} and $\hat{\theta}$ is
defined as for Laplace or delta method inference.\footnote{We note
  again that \myeq{monitor} is not the function we are optimizing.
  Even the simpler Laplace approximation is not clearly minimizing a
  well-defined distance function between the approximate Gaussian and
  true posterior~\citep{MacKay:1992}. Thus, while this approach is an
  approximate coordinate ascent algorithm, clearly characterizing the
  corresponding objective function is an open problem.}

\begin{figure*}[t]
\begin{center}
\centerline{\includegraphics[width=.96\textwidth]{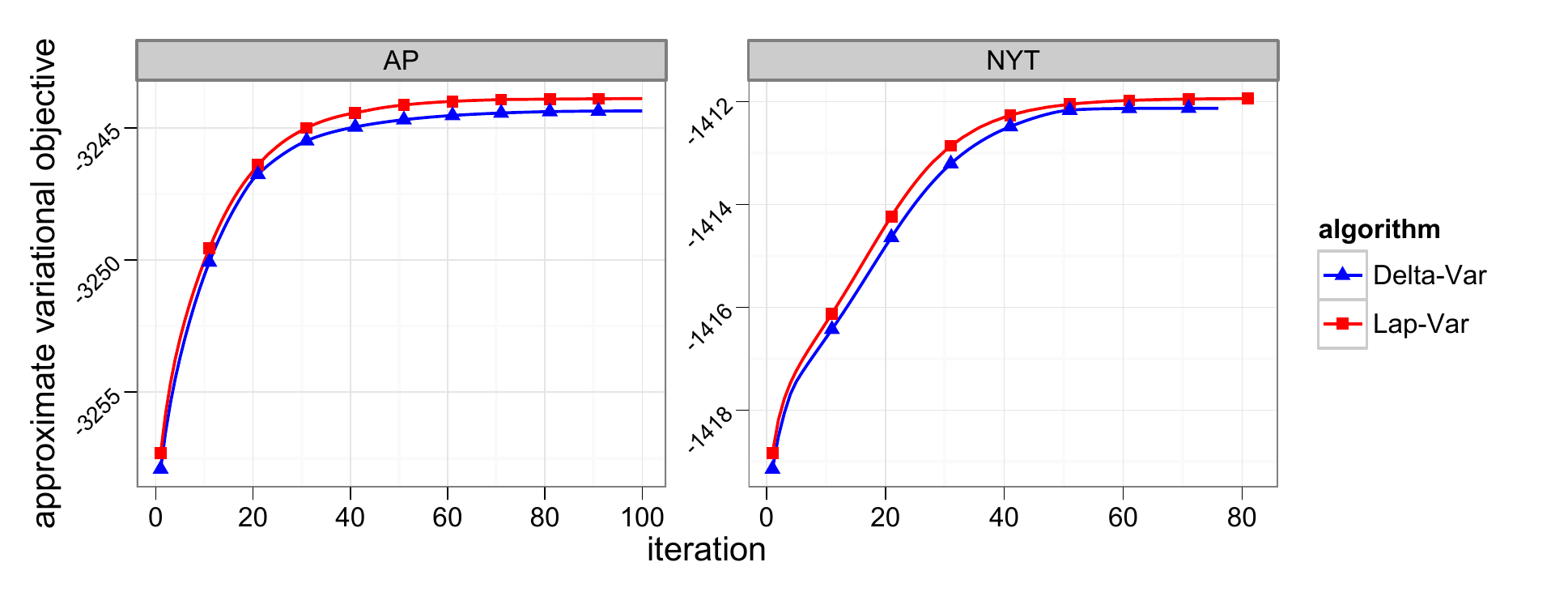}}
\caption{The approximate variational objective from \myeq{monitor}
  goes up as a function of the iteration.  This is for document-level
  inference in the correlated topic model.  The left plot is for a
  collection from the \textit{Associated Press}; the right plot is for
  a collection from the \textit{New York Times}. (See \mysec{ctm} and
  \mysec{ctm-study} for details about the model and data.)}
\label{fig:convergence}
\end{center}
\end{figure*}

\myfig{convergence} shows this score at each iteration for two runs of
inference in the correlated topic model. (See \mysec{ctm} for details
about the model.)  The approximate objective increases as the
algorithm proceeds, and these plots were typical.  In practice, as did
\citet{Braun:2010} in their setting, we found that this is a good
score to monitor.

\section{Example Models}\label{sec:example}

We have described a generic algorithm for approximate posterior
inference in nonconjugate models.  In this section we derive this
algorithm for several nonconjugate models from the research
literature: the correlated topic model~\citep{Blei:2007}, Bayesian
logistic regression~\citep{Jaakkola:1997a}, and hierarchical Bayesian
logistic regression~\citep{Gelman:2007}.  For each model, we identify
the variables---the nonconjugate variable $\theta$, conjugate variable
$z$, and observations $x$---and we calculate $f(\theta)$ from
\myeq{f-laplace}.  (The calculations of $f(\theta)$ are in the
appendices.)  In the next section, we study how our algorithms
perform when analyzing data under these models.\footnote{Python
implementations of our algorithms are available at
\url{http://www.cs.cmu.edu/~chongw/software/nonconjugate_inference.tar.gz}.}

\subsection{The Correlated Topic Model}\label{sec:ctm}

\begin{figure}[t]
 \begin{center}
   \centerline{\includegraphics[scale=0.5]{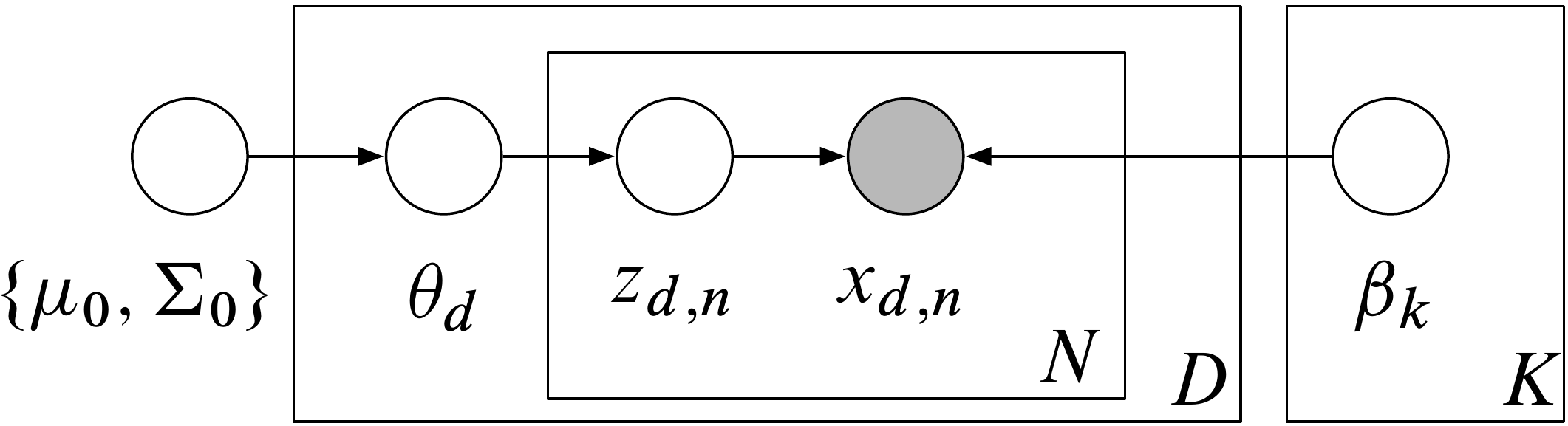}}
   \caption{The graphical representation of the correlated topic model
     (CTM).  The nonconjugate variable is $\theta$; the conjugate
     variable is the collection $z = z_{1:N}$; the observation is the
     collection of words $x = x_{1:N}$.}
   \label{fig:ctm}
 \end{center}
\end{figure}

Probabilistic topic models are models of document collections.  Each
document is treated as a group of observed words that are drawn from a
mixture model. The mixture components, called ``topics,'' are
distributions over terms that are shared for the whole collection;
each document exhibits them with individualized proportions.

Conditioned on a corpus of documents, the posterior topics place high
probabilities on words that are associated under a single theme; for
example, one topic may contain words like ``bat,'' ``ball,'' and
``pitcher.''  The posterior topic proportions reflect how each
document exhibits those themes; for example, a document may combine
the topics of \textit{sports} and \textit{health}.  This posterior
decomposition of a collection can be used for summarization,
visualization, or forming predictions about a document.
See~\cite{Blei:2012} for a review of topic modeling.

The per-document topic proportions are a latent variable.  In latent
Dirichlet allocation (LDA)~\citep{Blei:2003b}---which is the simplest
topic model---these are given a Dirichlet prior, which makes the model
conditionally conjugate.  Here we will study the correlated topic
model (CTM)~\citep{Blei:2007}.  The CTM extends LDA by replacing the
Dirichlet prior on the topic proportions with a logistic normal
prior~\citep{Aitchison:1982}.  This is a richer prior that can capture
correlations between occurrences of the components.  For example, a
document about \textit{sports} is more likely to also be about
\textit{health}.  The CTM is not conditionally conjugate.  But it is a
more expressive model: it gives a better fit to texts and provides new
kinds of exploratory structure.

Suppose there are $K$ topic parameters $\beta_{1:K}$, each of which is
a distribution over $V$ terms. Let $\pi(\theta)$ denote the
multinomial logistic function, which maps a real-valued vector to a
point on the simplex with the same dimension, $\pi(\theta) \propto
\exp\{\theta\}$.  The CTM assumes a document is drawn as follows:
\begin{enumerate*}
\item Draw log topic proportions $\theta \sim \mathcal{N}(\mu_0,
  \Sigma_0)$.
\item For each word $n$:
  \begin{enumerate*}
  \item Draw topic assignment $z_n \g \theta \sim {\rm Mult}(\pi(\theta))$.
  \item Draw word $x_n \g z_n, \beta \sim {\rm Mult}(\beta_{z_n})$.
  \end{enumerate*}
\end{enumerate*}
\myfig{ctm} shows the graphical model.  The topic proportions
$\pi(\theta)$ are drawn from a logistic normal distribution; their
correlation structure is captured in its covariance matrix $\Sigma_0$.
The topic assignment variable $z_n$ indicates from which topic the
$n$th word is drawn.

Holding the topics $\beta_{1:K}$ fixed, the main inference problem in
the CTM is to infer the conditional distribution of the document-level
hidden variables $p(\theta, z_{1:N} \g x_{1:N}, \beta_{1:K})$.  This
calculation is important in two contexts: it is used when forming
predictions about new data; and it is used as a subroutine in the
variational expectation maximization algorithm for fitting the topics
and logistic normal parameters (mean $\mu_0$ and covariance
$\Sigma_0$) with maximum likelihood.  The corresponding per-document
inference problem is straightforward to solve in LDA, thanks to
conditional conjugacy.  In the CTM, however, it is difficult because
the logistic normal on $\theta$ is not conjugate to the multinomial on
$z$.  \citet{Blei:2007} used a Taylor approximation designed
specifically for this model.  Here we apply the generic algorithm from
\mysec{inference}.

In terms of the earlier notation, the nonconjugate variable is the
topic proportions $\theta$, the conjugate variable is the collection
of topic assignments $z = z_{1:N}$, and the observation is the
collection of words $x = x_{1:N}$.  The variational distribution for
the topic proportions $\theta$ is Gaussian, $q(\theta) =
\mathcal{N}(\mu, \Sigma)$; the variational distribution for the topic
assignments is discrete, $q(z) = \prod_{n} q(z_n \g \phi_n)$ where
each $\phi_n$ is a distribution over $K$ elements.  In delta method
inference, as in~\cite{Braun:2010}, we restrict the variational
covariance $\Sigma$ to be diagonal to simplify the derivative of
\myeq{lb-II}. Laplace variational inference does not require this
simplification.  Appendix B gives the detailed derivations of the
algorithm.

Besides the CTM, this approach can be adapted to a variety of
nonconjugate topic models, including the topic evolution
model~\citep{Xing:2005}, Dirichlet-multinomial regression
~\citep{Mimno:2008}, dynamic topic
models~\citep{Blei:2006d,Wang:2008}, and the discrete infinite
logistic normal distribution~\citep{Paisley:2012b}.


\subsection{Bayesian Logistic Regression}
\label{sec:logistic-regression}

\begin{figure}[t]
 \begin{center}
   \centerline{\includegraphics[scale=0.5]{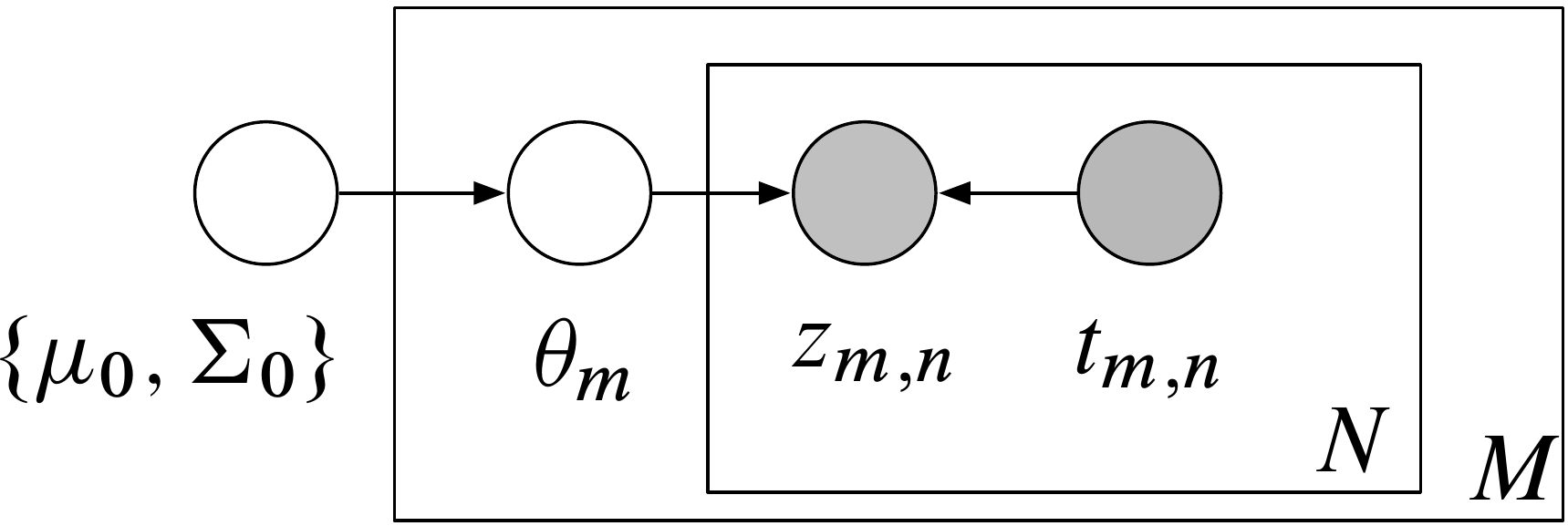}}
   \caption{The graphical representation of hierarchical logistic
     regression.  (When $M=1$, this is standard Bayesian logistic
     regression.)The nonconjugate variable is the vector of
     coefficients $\theta_m$, the conjugate variable is the collection
     of observed classes for each data point, $z_m = z_{m,1:N}$.  (In this
     case there is no additional observation $x$ downstream.)}
   \label{fig:bayesian-lg}
 \end{center}
\end{figure}

Bayesian logistic regression is a well-studied model for binary
classification~\citep{Jaakkola:1997a}.  It places a Gaussian prior on a
set of coefficients and draws class labels, conditioned on covariates,
from the corresponding logistic. Let $t_n$ is be a $p$-dimensional
observed covariate vector for the $n$th sample and $z_n$ be its class
label (an indicator vector of length two).  Let $\theta$ be the
real-valued coefficients in $\mathbb{R}^p$; there is a coefficient for
each feature.  Bayesian logistic regression assumes the following
conditional process:
\begin{enumerate*}
\item Draw coefficients $\theta \sim \mathcal{N} (\mu_0, \Sigma_0)$.
\item For each data point $n$ and its covariates $t_n$, draw its class
  label from
  \begin{equation*}
    z_n \g \theta, t_n \sim \textrm{Bernoulli}
    \left(\sigma(\theta^\top t_n)^{z_{n,1}}
      \sigma(-\theta^\top t_n)^{z_{n,2}}\right),
  \end{equation*}
  where $\sigma(y) \triangleq 1/\left(1+\exp(-y)\right)$ is the logistic
  function.
\end{enumerate*}
\myfig{bayesian-lg} shows the graphical model.  Given a dataset of
labeled feature vectors, the posterior inference problem is to compute
the conditional distribution of the coefficients $p(\theta \g z_{1:N},
t_{1:N})$.  The issue is that the Gaussian prior on the coefficients
is not conjugate to the conditional likelihood of the label.

This is a subset of the model class in \mysec{condition}.  The
nonconjugate variable $\theta$ is identical and the variable $z$ is
the collection of observed classes of each data point, $z_{1:N}$.
Note there is no additional observed variable $x$ downstream.  The
variational distribution need only be defined for the coefficients,
$q(\theta) = \mathcal{N}(\mu, \Sigma)$.  Using Laplace variational
inference, our approach recovers the standard Laplace approximation
for Bayesian logistic regression~\citep{Bishop:2006}.  This gives a
connection between standard Laplace approximation and
variational inference.  Delta method variational inference provides an
alternative.  Appendix C gives the detailed derivations.

An important extension of Bayesian logistic regression is hierarchical
Bayesian logistic regression~\citep{Gelman:2007}.  It simultaneously
models related logistic regression problems, and estimates the
hyperparameters of the shared prior on the coefficients.  With $M$
related problems, we construct the following hierarchical model:
\begin{enumerate*}
\item Draw the global hyperparameters,
  \begin{align}
    \Sigma_0^{-1} & \sim {\rm Wishart}(\nu,
    \Phi_0) \label{eq:multi-lgr-prior} \\
    \mu_0 &\sim \mathcal{N}(0, \Phi_1) \label{eq:multi-lgr-prior1}
  \end{align}
\item For each problem $m$:
  \begin{enumerate*}
  \item Draw coefficients $\theta_m \sim \mathcal{N} (\mu_0,
    \Sigma_0)$.
  \item For each data point $n$ and its covariates $t_{mn}$, draw its
    class label,
    \begin{equation*}
      z_{mn} \g \theta_m, t_{mn} \sim
      \textrm{Bernoulli}(\sigma(\theta_m^\top t_{mn})^{z_{mn,1}}
      \sigma(-\theta_m^\top t_{mn})^{z_{mn,2}}).
    \end{equation*}
  \end{enumerate*}
\end{enumerate*}

As for the CTM, we use nonconjugate inference as a subroutine in a
variational EM algorithm (where the M step is regularized). We
construct $f(\theta_m)$ in \myeq{f-laplace} separately for each
problem $m$, and fit the hyperparameters $\mu_0$ and $\Sigma_0$ from
their approximate expected sufficient statistics~\citep{Bishop:2006}.
This amounts to MAP estimation with priors as specified above. See
Appendix C for the complete derivation.

Finally, we note that logistic regression is a generalized linear
model with a binary response and canonical link
function~\citep{McCullagh:1989}.  It is straightforward to use our
algorithms with other Bayesian generalized linear models (and their
hierarchical forms).

\section{Empirical Study}
\label{sec:experiments}

We studied nonconjugate variational inference with correlated topic
models and Bayesian logistic regression.  We found that nonconjugate
inference is more accurate than the existing methods tailored to
specific models.  Between the two nonconjugate inference algorithms,
we found that Laplace inference is faster and more accurate than delta
method inference.

\subsection{The Correlated Topic Model}
\label{sec:ctm-study}

We studied Laplace inference and delta method inference in the CTM.
We compared it to the original inference algorithm of
\citet{Blei:2007}.

We analyzed two collections of documents.  The \textit{Associated
Press} (AP) collection contains 2,246 documents from the {\it
Associated Press}.  We used a vocabulary of 10,473 terms, which
gave a total of 436K observed words.  The {\it New York Times}
(NYT) collection contains 9,238 documents from the {\it New York
Times}.  We used a vocabulary of 10,760 terms, which gave a total
of 2.3 million observed words.  For each corpus we used 80\% of
the documents to fit models and reserved 20\% to test them.

We fitted the models with variational EM.  At each iteration, the
algorithm has a set of topics $\beta_{1:K}$ and parameters to the
logistic normal $\{\mu_0, \Sigma_0\}$.  In the E-step we perform
approximate posterior inference with each document, estimating its
topic proportions and topic assignments.  In the M-step, we
re-estimate the topics and logistic normal parameters.  We fit models
with different kinds of E-steps, using both of the nonconjugate
inference methods from \mysec{inference} and the original approach
of~\citet{Blei:2007}. To initialize nonconjugate inference we set the
variational mean parameter $\mu=0$ for log topic proportions $\theta$
and computed the corresponding updates for the topic assignments $z$.
We initialize the topics in variational EM to random draws from a
uniform Dirichlet.

With nonconjugate inference in the E-step, variational EM
approximately optimizes a bound on the marginal probability of the
observed data. We can calculate an approximation of this bound with
\myeq{monitor} summed over all documents.  We monitor this quantity as
we run variational EM.

To test our fitted models, we measured predictive performance on
held-out data with predictive distributions derived from the posterior
approximations.  We follow the testing framework
of~\citet{Asuncion:2009} and~\citet{Blei:2007}.  We fix fitted topics
and logistic normal parameters $M = \{\beta_{1:K}, \mu_0, \Sigma_0\}$.
We split each held-out document in to two halves $(\bm w_{1}, \bm
w_{2})$ and form the approximate posterior log topic proportions
$q_{\bm w_{1}}(\theta)$ using one of the approximate inference
algorithms and the first half of the document $\bm w_1$.  We use
this to form an approximate predictive distribution,
\begin{eqnarray*}
  p(w \g \bm w_{1}, M) &\approx&
  \textstyle \int_{\theta} \sum_{z} p(w \g z, \beta_{1:K}) q_{\bm w_1}(\theta)
  d\theta \\
  &\approx& \textstyle \sum_{k=1}^{K} \beta_{kw} \pi_k,
\end{eqnarray*} where $\pi_k \propto
\exp\{\expect{q}{\theta_k}\}$.  Finally, we evaluate the log
probability of the second half of the document
using that predictive distribution; this is the {\it heldout
  log likelihood}.  A better model and inference method will give
higher predictive probabilities of the unseen words.  Note that
this testing framework puts the approximate posterior
distributions on the same playing field.  The quantities are
comparable regardless of how the approximate posterior is formed.

\begin{figure*}[t]
\begin{center}
  \centerline{\includegraphics[width=1.0\textwidth]{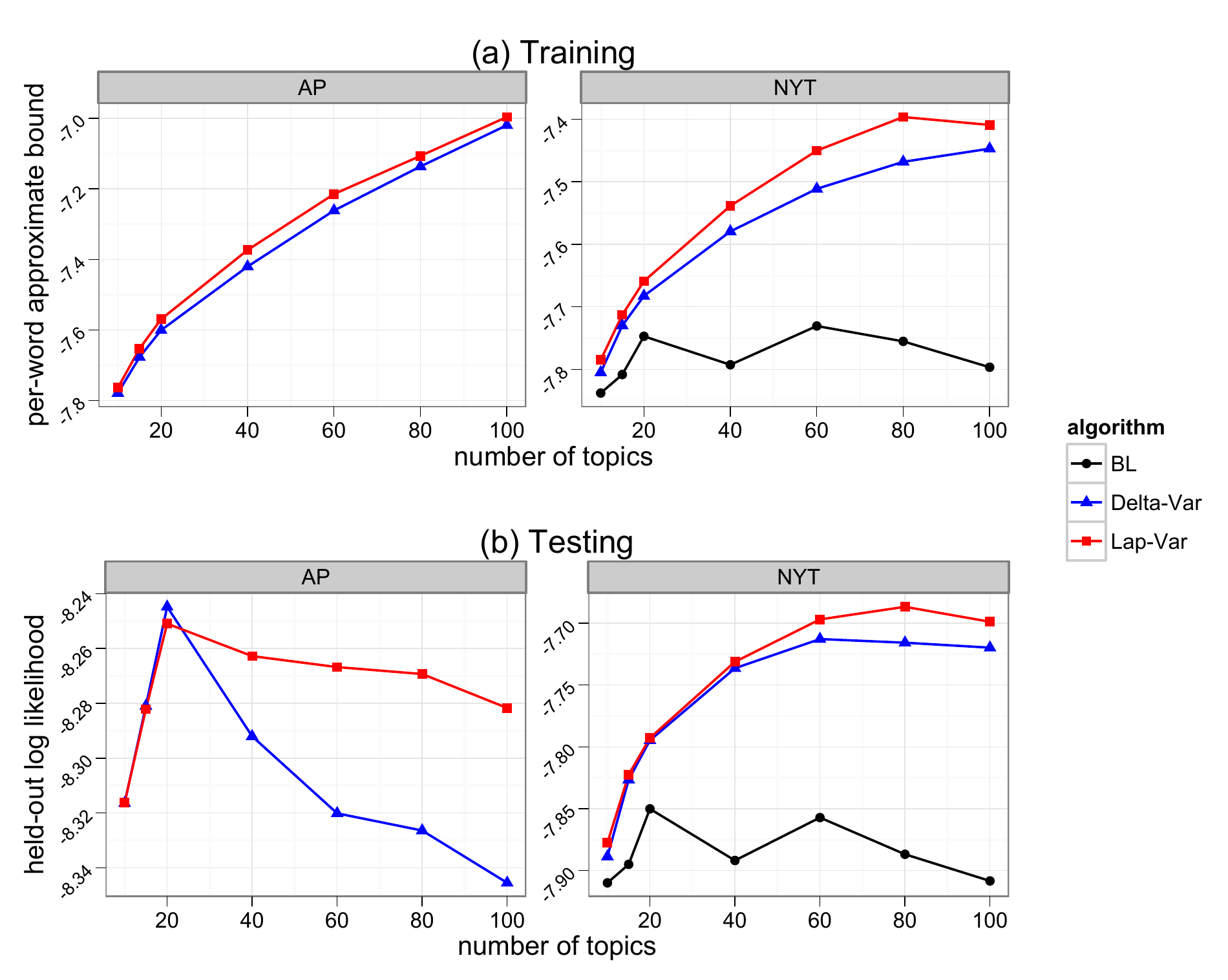}}
  \caption{  Laplace variational inference is ``Lap-Var''; delta
    method variational inference is ``Delta-Var''; Blei and
    Lafferty's method is ``BL.''  (a) Approximate per-word lower
    bound against the number of topics. A good approximation will
    go up as the number of topics increases, but not necessarily
    indicate a better predictive performance on the heldout data.
    (b) Per-word held-out log likelihood against the number of
    topics.  Higher numbers are better.  Both nonconjugate
    methods perform better than Blei and Lafferty's method.
    Laplace inference performs best. Blei and Lafferty's method
    was erratic in both collections.  (It is not plotted for the
    AP collection.)
  }
\label{fig:train-test-likelihood}
\end{center}
\end{figure*}

\myfig{train-test-likelihood} shows the per-word approximate bound and
the per-word heldout likelihood as functions of the number of
topics. \myfig{train-test-likelihood} (a) indicates that the
approximate bounds from nonconjugate inference generally go up as the
number of topics increases.  This is a property of a good
approximation because the marginal certainly goes up as the number of
parameters increases.  In contrast, Blei and Lafferty's (2007)
objective (which is a true bound on the marginal of the data) behaves
erratically. This is illustrated for the \textit{New York Times}
corpus; on the {\it Associated Press} corpus, it does not come close
to the approximate bound and is not plotted.

\myfig{train-test-likelihood} (b) shows that on heldout data, Blei and
Lafferty's approach, tailored for this model, performed worse than
both of our algorithms.  Our conjecture is that while this method
gives a strict lower bound on the marginal, it might be a loose bound
and give poor predictive distributions.  Our methods use an
approximation which, while not a bound, might be closer to the
objective and give better predictive distributions.  The heldout
likelihood plots also show that when the number of topics increases
the algorithms eventually overfit the data.  Finally, note that
Laplace variational inference was always better than both other
algorithms.

\begin{figure*}[t]
\begin{center}
\centerline{\includegraphics[width=1.\textwidth]{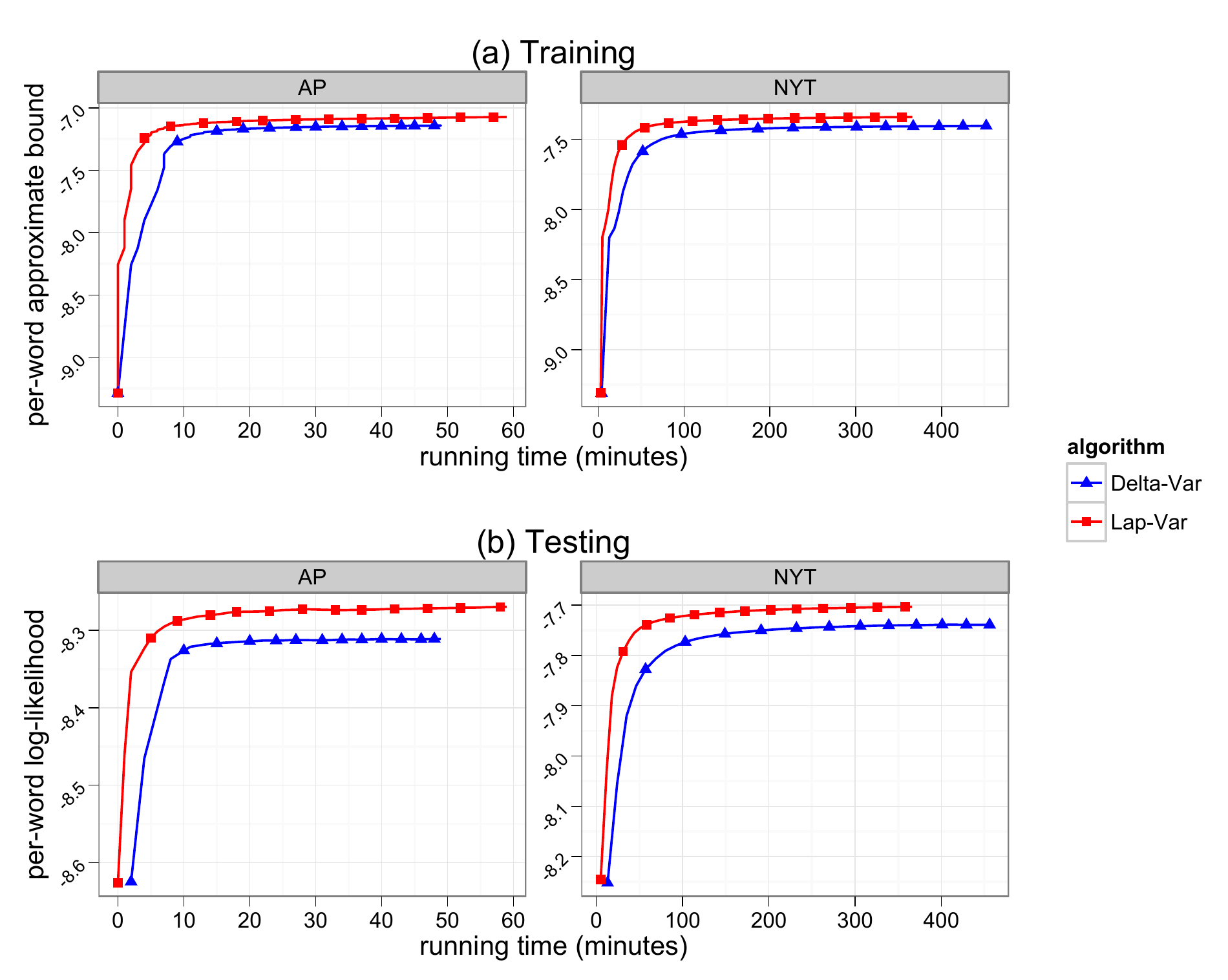}}
\caption{In this figure, we set the number of topics as $K=60$.
(Others are similar.) (a) The per-word approximate bound during
model fitting with variational EM.  Though it is an
approximation of the variational EM objective, it converges in
practice. (b) The per-word heldout likelihood during the model
fitting with variational EM.  Laplace inference performs best in
terms of speed and predictive performance.}
\label{fig:train-test-time}
\end{center}
\end{figure*}

Finally, \myfig{train-test-time} shows the approximate bound and the
heldout log likelihood as functions of running time.\footnote{We did
  not formally compare the running time of \cite{Blei:2007}'s method
  because we used the authors' C implementation, while ours is in
  Python. We observed that their method took more than five times
  longer than ours.} From \myfig{train-test-time} (a), we see that
even though variational EM is not formally optimizing this approximate
objective (see \myeq{monitor}), the increase at each iteration
suggests that the marginal probability is also increasing.  The plot
also shows that Laplace inference converges faster than delta method
inference. \myfig{train-test-time} (b) confirms that Laplace inference
is both faster and gives better predictive performance.

\subsection{Bayesian Logistic Regression}

We studied our algorithms on Bayesian logistic regression in both
standard and hierarchical settings. In the standard setting, we
analyzed two datasets. With the {\it Yeast}
data~\citep{Elisseeff:2001}, we form a predictor of gene functional
classes from features composed of micro-array expression data and
phylogenetic profiles. The dataset has 1,500 genes in the training set
and 917 genes in the test set.  For each gene there are 103 covariates
and up to 14 different gene functional classes (14 labels).  This
corresponds to 14 independent binary classification problems.  With
the {\it Scene} data~\citep{Boutell:2004}, we form a predictor of
scene labels from image features.  It contains 1,211 images in the
training set and 1,196 images in the test set. There are 294 images
features and up to 6 scene labels per image.  This corresponds to 6
independent binary classification problems.\footnote{The {\it Yeast}
  and {\it Scene} data are at
  http://mulan.sourceforge.net/datasets.html.}

We used two performance measures. First we measured accuracy, which is
the proportion of test-case examples correctly labeled. Second, we
measured average log predictive likelihood.  Given a test-case input
$t$ with label $z$, we compute the log predictive likelihood,
\begin{equation*}
  \log p(z \g \mu, t) =  z_1\log \sigma(\mu^\top t) + z_2 \log
  \sigma(-\mu^\top t),
\end{equation*}
where $\mu$ is the mean of variational distribution $q(\theta) =
\mathcal{N}(\mu, \Sigma)$. Higher likelihoods indicate a better fit.
For both accuracy and predictive likelihood, we used cross validation
to estimate the generalization performance of each inference
algorithm.  We set the priors $\mu_0 = 0$ and $\Sigma_0 = I$.

We compared Laplace inference (\mysec{embed-laplace}), delta method
inference (\mysec{lower-bound}), and the method
of~\cite{Jaakkola:1997a}.  Jaakkola and Jordan's (1996) method
preserves a lower bound on the marginal likelihood with a first-order
Taylor approximation and was developed specifically for Bayesian
logistic regression.  (We note that Blei and Lafferty's
bound-preserving method for the CTM was built on this technique.)

Table~\ref{tab:standard-task} gives the results.  To compare methods
we compute the difference in score (accuracy or log likelihood) on the
independent binary classification problems, and then perform a
standard t-test (at level $0.05$) to test if the mean of the
differences is larger than 0.  Laplace inference and delta method
inference gave slightly better accuracy than Jaakkola and Jordan's
method, and much better log predictive likelihood.\footnote{Previous
  literature, e.g., .  \citet{Xue:2007} and \citet{Archambeau:2011}
  treat {\it Yeast} and {\it Scene} as multi-task problems.  In our
  study, we found that our standard Bayesian logistic regression
  algorithms performed the same as the algorithms developed in these
  papers.}  The t-test showed that both Laplace and delta method
inference are better than Jaakkola and Jordan's method.


We next examined a dataset of student performance in a collection of
schools.  With the {\it School} data, our goal is to use various
features of a student to predict whether he or she will perform above
or below the median on a standardized exam.\footnote{The data is
  available at http://multilevel.ioe.ac.uk/intro/datasets.html.}  The
data came from the Inner London Education Authority.  It contains
examination records from 139 secondary schools for the years 1985,
1986 and 1987.  It is a random 50\% sample with 15,362 students.  The
students' features contain four student-dependent features and
school-dependent features.  The student dependent features are the
year of the exam, gender, VR band (individual prior attainment data),
and ethnic group; the school-dependent features are the percentage of
students eligible for free school meals, percentage of students in VR
band 1, school gender, and school denomination.  We coded the binary
indicator of whether each was below the median (``bad'') or above
(``good'').  We use the same 10 random splits of the data
as~\cite{Argyriou:2008}.


In this data, we can either treat each school as a separate
classification problem, pool all the schools together as a single
classification problem, or analyze them with hierarchical logistic
regression (\mysec{logistic-regression}).  The hierarchical model
allows the predictors for each school to deviate from each other, but
shares statistical strength across them.
Let $p$ be the number of covariates.  We set the prior on the
hyperparameters to the coefficients to $\nu = p+100$,
$\Phi_0=0.01I$, and $\Phi_1 = 0.01I$ ( see \myeq{multi-lgr-prior}
and \myeq{multi-lgr-prior1}) to favor sparsity.  We initialized
the variational distributions to $q(\theta)=\mathcal{N}(0, I)$.

Table~\ref{tab:multi-task} gives the results.  A standard t-test (at
level $0.05$) showed that the hierarchical models are better than the
non-hierarchical models both in terms of accuracy and predictive
likelihood.  With predictive likelihood, Laplace variational inference
in the hierarchical model is significantly better than all other
approaches.



\begin{table*}[t]
\begin{center}
\begin{tabular}{lcc|cc}
  \hline
  & \multicolumn{2}{c}{\textit{Yeast}} &
  \multicolumn{2}{c}{\textit{Scene}} \\

  & Accuracy & Log Likelihood & Accuracy & Log
  Likelihood \\
  \hline
  Jaakkola and Jordan (1996) & 79.7\% & -0.678 & 87.4\% & -0.670 \\
  Laplace inference &  \textbf{80.1\%} & {\bf -0.449} & \textbf{89.4\%} & {\bf -0.259}  \\
  Delta method inference & {\bf 80.2\%} & {\bf -0.450} & {\bf 89.5\%} & -0.265  \\
  \hline
\end{tabular}
\end{center}
\caption{Comparison of the different methods for Bayesian logistic
  regression using accuracy and averaged log predictive likelihood.
  Higher numbers are better.  These results are averaged from five
  random starts.  (The variance is too small to report.)  Bold results
  indicate significantly better performance using a standard
  t-test. Laplace and delta method inference perform best.}
  \label{tab:standard-task}
\end{table*}

\begin{table*}[t]
\begin{center}
\begin{tabular}{ll|cc}
  \hline
  && Accuracy & Log Likelihood  \\
  \hline
  \textit{Separate} &&&\\
  & Jaakkola and Jordan (1996) & 70.5\% & -0.684  \\
  & Laplace inference &  70.8\% & -0.569 \\
  & Delta inference  & 70.8\%& -0.571  \\
  \textit{Pooled} &&& \\
  & Jaakkola and Jordan (1996) & 71.2\%  & -0.685   \\
  & Laplace inference & 71.3\% & -0.557   \\
  & Delta inference  & 71.3\% & -0.557 \\
  \textit{Hierarchical} &&& \\
  & Jaakkola and Jordan (1996) & 71.3\%  & -0.685   \\
  & Laplace inference & {\bf 71.9\%}& {\bf -0.549}   \\
  & Delta inference  & {\bf 71.9\%} & -0.559  \\
  \hline
\end{tabular}
\end{center}

\caption{Comparison of the different methods on the \textit{School}
  data using accuracy and averaged log predictive likelihood.  Results
  are averaged from 10 random splits.  (The variance is too small to
  report.)  We compared Laplace inference, delta inference and
  Jaakkola and Jordan's (1996) method in three settings: separate
  logistic regression models for each school, a pooled logistic
  regression model for all schools, and the hierarchical logistic
  regression model in \mysec{logistic-regression}.  Bold indicates
  significantly better performance by a standard t-test (at level 0.05).  The
  hierarchical model performs best. \label{tab:multi-task}}
\end{table*}

\section{Discussion}
\label{sec:discussion}

We developed Laplace and delta method variational inference, two
strategies for variational inference in a large class of nonconjugate
models.  These methods approximate the variational objective function
with a Taylor approximation, each in a different way. We studied them
in two nonconjugate models and showed that they work well in practice,
forming approximate posteriors that lead to good predictions. In the
examples we analyzed, our methods worked better than methods tailored
for the specific models at hand. Between the two, Laplace inference
was better and faster than delta method inference.  These methods
expand the scope of variational inference.





\section*{Appendix A: Generalization to Complex Models}
We describe how we can generalize our approaches to more complex
models. Suppose we have a directed probabilistic model with
latent variables $\theta = \theta_{1:m}$ and observations $x$.
(We will not differentiate notation between conjugate and
nonconjugate variables.) The log joint likelihood of all latent
and observed variables is
\begin{equation*}
  \log p(\theta, x) =
  \sum_{i=1}^{m} \log p(\theta_i \g \theta_{\pi_i}) + \log p(x \g \theta),
\end{equation*}
where $\pi_i$ are the indices of the parents of $\theta_i$, the
variables it depends on.

Our goal is to approximate the posterior distribution $p(\theta \g
x)$. Similar to the main paper, we use mean-field variational
inference~\citep{Jordan:1999}.  We posit a fully-factorized
variational family
$$q(\theta) = \prod_{i=1}^{m} q(\theta_i)$$
and optimize the each factor $q(\theta_i)$ to find the member closest
in KL-divergence to the posterior.

As in the main paper, we solve this optimization problem with
coordinate ascent, iteratively optimizing each variational factor
while holding the others fixed. Recall that \cite{Bishop:2006} shows
that this leads to the following update
\begin{equation}
  q(\theta_i) \propto \exp\left\{\E{-i}{\log p(\theta,x)}\right\},
  \label{eq:bishop-update}
\end{equation}
where $\E{-i}{\cdot}$ denotes the expectation with respect to
$\prod_{j,j\neq i}q(\theta_j)$.

Many of the terms of the log joint will be constant with respect to
$\theta_i$ and absorbed into the constant of proportionality. This
allows us to simplify the update in~\myeq{bishop-update} to be
$q(\theta_i) \propto \exp\left\{f(\theta_i)\right\}$ where
\begin{equation}
  f(\theta_i) =
  \E{-i}{\log p(\theta_i \g \theta_{\pi_i})} +
  \sum_{\{j: i \in \pi_j\}} \E{-i}{\log p(\theta_j \g \theta_{\pi_j})} +
  \E{-i}{\log p(x \g \theta)}.
  \label{eq:general-laplace-f}
\end{equation}
As in the main paper, this update is not tractable in general.  We use
Laplace variational inference (\mysec{embed-laplace}) to approximate
it, although delta method variational inference (\mysec{lower-bound})
is also applicable. In Laplace variational inference, we take a Taylor
approximation of $f(\theta_i)$ around its maximum $\hat{\theta}_i$.
This naturally leads to $q(\theta_i)$ as a Gaussian factor,
\begin{equation*}
  q^*(\theta_i) \approx {\cal N}(\hat{\theta}_i, -\triangledown^2
  f(\hat{\theta}_i)^{-1}).
\end{equation*}
The main paper considers the case where $\theta$ is a single random
variable and updates its variational distribution. In the more general
coordinate ascent setting considered here, we need to compute or
approximate the expected log probabilities (and their derivatives) in
\myeq{general-laplace-f}.

Now suppose each factor is in the exponential family. (This is weaker
than the conjugacy assumption, and describes most graphical models
from the literature.) The log joint likelihood becomes
\begin{equation}
  \log p(\theta, x) =
  \sum_{i=1}^{m} \left(
    \eta(\theta_{\pi_i})^\top t(\theta_i) -
    a(\eta(\theta_{\pi_i}))
  \right) +
  \log p(x \g \theta), \label{eq:log-joint-appendix}
\end{equation}
where $\eta(\cdot)$ are natural parameters, $t(\cdot)$ are sufficient
statistics, and $a(\eta(\cdot))$ are log normalizers.  (All are
overloaded.)  Substituting the exponential family assumptions into
$f(\theta_i)$ gives
\begin{align*}
  f(\theta_i) &=  \E{-i}{\eta(\theta_{\pi_i})}^\top t(\theta_i) \nonumber \\
  & + \textstyle \sum_{\{j: i \in \pi_j\}} \left( \E{-i}{\eta(\theta_{\pi_j})}^\top
    \E{-i}{t(\theta_j)} - \E{-i}{a(\eta(\theta_{\pi_j}))} \right) \\
  & + \E{-i}{t(\theta)}^\top t(x) - \E{-i}{a(\eta(\theta))}. \nonumber
\end{align*}
Here we can use further Taylor approximations of the natural
parameters $\eta(\cdot)$, sufficient statistics $t(\cdot)$, and log
normalizers $a(\cdot)$ in order to easily take their expectations.

Finally, for some variables we may be able to exactly compute
$f(\theta_i)$ and form the $q^*(\theta_i)$ without further
approximations.  (These are conjugate variables for which the complete
conditional $p(\theta_i \g \theta_{-i}, x)$ is available in closed
form.) These variables were separated out in the main paper; here we
note that they can be updated exactly in the coordinate ascent
algorithm.

\section*{Appendix B: The Correlated Topic Model}
The correlated topic model is described in \mysec{ctm}.  We identify
the quantities from~\myeq{z-dist} and~\myeq{x-dist} that we need to
compute $f(\theta)$ in~\myeq{f-laplace},
\begin{align*}
  h(z) &=1, \ \ t(z) = \textstyle \sum_n z_n, \\
  \eta(\theta) &= \textstyle
  \theta-\log\left\{\sum_k\exp\{\theta_k\}\right\} \\
  a(\eta(\theta)) &= 0.
\end{align*}
With this notation,
\begin{align*}
  f(\theta) = \textstyle \eta(\theta)^\top\expect{q(z)}{t(z)} -\frac{1}{2}
  (\theta-\mu_0)^\top\Sigma_0^{-1}(\theta - \mu_0),
\end{align*}
where $\expect{q(z)}{t(z)}$ is the expected word counts of each topic under the
variational distribution $q(z)$.

Let $\pi = \eta(\theta)$ be the topic proportions.  Using ${\partial
  \pi_i }/ {\partial \theta_j} = \pi_i(1_{[i=j]} - \pi_j)$, we obtain
the gradient and Hessian of the function $f(\theta)$ in the CTM,
\begin{align*}
  \triangledown f(\theta) &= \textstyle \expect{q(z)}{t(z)} - \pi
  \sum_{k=1}^K\left[\expect{q(z)}{t(z)} \right]_k -\Sigma_0^{-1}(\theta - \mu_0), \\
  \triangledown^2 f(\theta)_{ij} &= \textstyle  (-\pi_i 1_{[i=j]} + \pi_i \pi_j)
  \sum_{k=1}^K\left[\expect{q(z)}{t(z)}\right]_k-(\Sigma_0^{-1})_{ij}.
\end{align*}
where $1_{[i=j]}=1$ if $i=j$ and $0$ otherwise. Note that
$\triangledown f(\theta)$ is all we need for Laplace inference.

In delta method variational inference, we also need to compute
the gradient of
\begin{align*}
  {\rm Trace}\left\{\triangledown^2 f(\theta) \Sigma\right\} &=
  \textstyle \left(-\sum_{k=1}^K \pi_k \Sigma_{kk} + \pi^T \Sigma
    \pi\right) \sum_{k=1}^K\left[\expect{q(z)}{t(z)}\right]_k - {\rm
    Trace}(\Sigma_0^{-1}\Sigma).
\end{align*}
Following~\citep{Braun:2010}, we assume $\Sigma$ is diagonal in the
delta method. (In Laplace inference, we do not need this assumption.)
This gives
\begin{align*}
  & \frac{\partial {\rm Trace}\left\{\triangledown^2 f(\theta)
      \Sigma\right\}} { \partial \theta_i} = \textstyle
  \pi_i(1-2\pi_i)(\sum_k \pi_k\Sigma_{kk} -1).
\end{align*}
These quantities let us implement the algorithm
in~\myfig{lp-algorithm} to infer the per-document posterior of the CTM
hidden variables.

As we discussed \mysec{ctm}, we use this algorithm in variational EM
for finding maximum likelihood estimates of the model parameters.  The
E-step runs posterior inference on each document.  Since the
variational family is the same, the M-step is as described in
~\citet{Blei:2007}.

\section*{Appendix C: Bayesian Logistic Regression}
Bayesian logistic regression is described in
\mysec{logistic-regression}.

The distribution of the observations $z_{1:N}$ fit into the
exponential family as follows,
\begin{align*}
  h(z) &=1, \ \ t(z) = \textstyle [z_1, \dots, z_N], \\ \eta(\theta) &
  = [\log\sigma(\theta^\top t_n), \log\sigma(-\theta^\top
  t_n)]_{n=1}^N \\ a(\eta(\theta)) &= 0.
\end{align*}
In this set up, $t(z)$ represents the whole set of labels. Since $z$ is
observed, its ``expectation'' is just itself. With this notation,
$f(\theta)$ from ~\myeq{f-laplace} is
\begin{align*}
  f(\theta) = \textstyle \eta(\theta)^\top t(z) -\frac{1}{2}
  (\theta-\mu_0)^\top\Sigma_0^{-1}(\theta - \mu_0).
\end{align*}
The gradient and Hessian of $f(\theta)$ are
\begin{align*}
  \triangledown f(\theta) &= \textstyle \sum_{n=1}^N t_n
  \left(z_{n,1}-\sigma(\theta^T t_n)\right) -\Sigma_0^{-1}(\theta -
  \mu_0), \\
  \triangledown^2 f(\theta) &= \textstyle -\sum_{n=1}^N\sigma(\theta^T
  t_n)\sigma(-\theta^T t_n) t_nt_n^T -\Sigma_0^{-1}.
\end{align*}
This is the standard Laplace approximation to Bayesian logistic
regression~\citep{Bishop:2006}.

For delta variational inference, we also need the gradient for ${\rm
Trace}\left\{\triangledown^2 f(\theta) \Sigma\right\}$.  It is
\begin{align*}
  & \frac{\partial {\rm Trace}\left\{\triangledown^2 f(\theta)
      \Sigma\right\}} { \partial \theta_i} =
  -\sum_{n=1}^N\sigma(\theta^T t_n)\sigma(-\theta^T t_n)
  (1-2\sigma(\theta^T t_n)) t_n t_n^T\Sigma t_n.
\end{align*}
Here we do not need to assume $\Sigma$ is diagonal.  This Hessian is
already diagonal.


\paragraph{Hierarchical logistic regression.} Here we describe how we
update the global hyperparameters $(\mu_0, \Sigma_0)$ (Eq.
\ref{eq:multi-lgr-prior} and \ref{eq:multi-lgr-prior1}) in
hierarchical logistic regression.  At each iteration, we first compute
the variational distribution of coefficients $\theta_m$ for each
problem $m = 1,..., M$,
\begin{align*}
  q(\theta_m) = \mathcal{N}(\mu_m, \Sigma_m).
\end{align*}
We then estimate the global hyperparameters $(\mu_0, \Sigma_0)$ using
the MAP estimate.  These come from the following update equations,
\begin{align*}
  \mu_0 & =\left( \frac{\Sigma_0\Phi_1^{-1}}{M} +
  I_p \right)^{-1}\frac{\sum_{m=1}^M \mu_m}{M},  \\
  \Sigma_0 & =\frac{\Phi_0^{-1} + \sum_{m=1}^M
(\mu_m - \mu_0)(\mu_m - \mu_0)^\top}{M + \nu - p -1},
\end{align*}
where $p$ is the dimension of coefficients $\theta_m$.

\paragraph{Acknowledgements.} We thank Jon McAuliffe and the anonymous
reviewers for their valuable comments.  Chong Wang was supported by
Google Ph.D. and Siebel Scholar Fellowships.  David M. Blei is
supported by NSF IIS-0745520, NSF IIS-1247664, NSF IIS-1009542, ONR
N00014-11-1-0651, and the Alfred P. Sloan foundation.

\vskip 0.2in
\bibliography{nonconjugate-inference-bib.bib}

\end{document}